\documentclass{article}

\usepackage{arxiv}

\usepackage[utf8]{inputenc} % allow utf-8 input
\usepackage[T1]{fontenc}    % use 8-bit T1 fonts
\usepackage{hyperref}       % hyperlinks
\usepackage{url}            % simple URL typesetting
\usepackage{booktabs}       % professional-quality tables
\usepackage{amsfonts}       % blackboard math symbols
\usepackage{nicefrac}       % compact symbols for 1/2, etc.
\usepackage{microtype}      % microtypography
\usepackage{lipsum}
\usepackage{graphicx}
\usepackage{amsmath}
\usepackage{amssymb}
\newcommand{\argminD}{\arg\!\min} % AlfC
\usepackage{tabularx}
\usepackage{siunitx}
\usepackage{wrapfig}
\newcommand{\beginsupplement}{%
        \setcounter{table}{0}
        \renewcommand{\thetable}{S\arabic{table}}%
        \setcounter{figure}{0}
        \renewcommand{\thefigure}{S\arabic{figure}}%
     }

\title{GANalyze: Toward Visual Definitions of Cognitive Image Properties}

\author{
  Lore~Goetschalckx*\\
  MIT, KU Leuven\\
  \texttt{lgoetsch@mit.edu} \\
   \And
 Alex~Andonian* \\
  MIT\\
  \texttt{aandonia@mit.edu} \\
  \And
  Aude~Oliva \\
  MIT\\
  \texttt{oliva@mit.edu} \\
  \And
  Phillip~Isola \\
  MIT\\
  \texttt{phillipi@mit.edu} 
}

\begin{document}
\maketitle

% \begin{figure}
%   \centering
%   \includegraphics[width=480pt]{latex/teaser2.pdf}
%   \caption{Sample figure caption.}
% \end{figure}

%\begin{multicols}{2}
\begin{figure}[h!]
\begin{minipage}{\textwidth}
	\centering
	\vspace{0.0in}

\includegraphics[width=480pt]{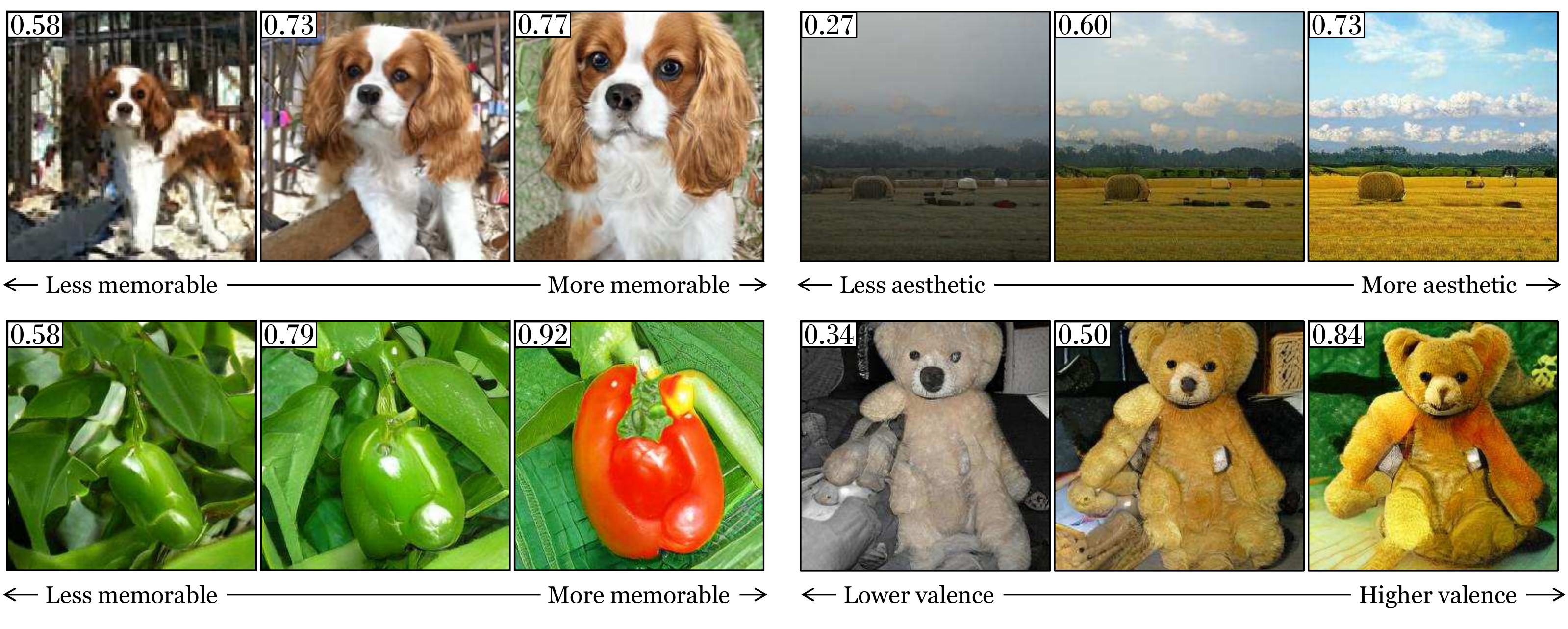}
%\hspace{10pt}
%\includegraphics[width=240pt]{latex/examples_teaser_1.pdf}
\caption{\textbf{Visualizations produced by the proposed GANalyze framework}. The middle columns represent generated images serving as the original seed. The originals are then modified to be characterized more (right) or less (left) by a given property of interest (memorability, aesthetics, or emotional valence). The images' respective property scores are presented in their top left corner.}

\label{fig:teaser}
\end{minipage}
%\vspace{-1.3in}

\end{figure} 

%\end{multicols}

\vspace{0.3in}
\begin{abstract}
        We introduce a framework that uses Generative Adversarial Networks (GANs) to study cognitive properties like memorability, aesthetics, and emotional valence. These attributes are of interest because we do not have a concrete visual definition of what they entail. What does it look like for a dog to be more or less memorable? GANs allow us to generate a manifold of natural-looking images with fine-grained differences in their visual attributes. By navigating this manifold in directions that increase memorability, we can visualize what it looks like for a particular generated image to become more or less memorable. The resulting ``visual definitions" surface image properties (like ``object size") that may underlie memorability. Through behavioral experiments, we verify that our method indeed discovers image manipulations that causally affect human memory performance. We further demonstrate that the same framework can be used to analyze image aesthetics and emotional valence. Visit the GANalyze website at \url{http://ganalyze.csail.mit.edu/}.

\end{abstract}

% keywords can be removed

%%%%%%%%%%%%%%%%%%%%%%%%%%%%%%%%%%%%%%%%%%%%%%%%%%%%%%%%%%%
\section{Introduction}
%%%%%%%%%%%%%%%%%%%%%%%%%%%%%%%%%%%%%%%%%%%%%%%%%%%%%%%%%%%

Why do we remember the things we do? Decades of work have provided numerous explanations: we remember things that are out of context \cite{zoya,standing}, that are emotionally salient \cite{emotions}, that involve people~\cite{Isola2014}, etc. But a picture is, as they say, worth a thousand words. What does it \emph{look like} to make an image more or less memorable? The same questions can be asked for many cognitive visual properties: what visual changes can take a bland foggy seascape and add just the right colors and tones to make it serenely beautiful.

Attributes like memorability, aesthetics, and emotional valence are of special interest because we do not have concrete definitions of what they entail. This contrasts with attributes like ``object size" and "smile". We know exactly what it means to zoom in on a photo, and it's easy to imagine what a face looks like as it forms a smile. It's an open question, on the other hand, what exactly do changes in ``memorability" look like? Previous work has built powerful predictive models of image memorability~\cite{Isola2014,Khosla_2015_ICCV} but these have fallen short of providing a fine-grained visual explanation of what underlies the predictions.

In this paper, we propose a new framework, GANalyze, based on Generative Adversarial Networks (GAN)~\cite{goodfellow}, to study the visual features and properties that underlie high-level cognitive attributes. We focus on image memorability as a case study, but also show that the same methods can be applied to study image aesthetics and emotional valence.

Our approach leverages the ability of GANs to generate a continuum of images with fine-grained differences in their visual attributes. We can learn how to navigate the GAN's latent space to produce images that have increasing or decreasing memorability, according to an off-the-shelf memorability predictor~\cite{Khosla_2015_ICCV}. Starting with a seed image, this produces a sequence of images of increasing and decreasing predicted memorability (see Figure 1). By showing this visualization for a diverse range of seed images, we come up with a catalog of different image sequences showcasing a variety of visual effects related to memorability. We call this catalog a \emph{visual definition} of image memorability. GANalyze thereby offers an alternative to the non-parametric approach in which real images are simply sorted on their memorability score to visualize what makes them memorable (example shown in Figure~\ref{fig:nonparametric}). The parametric, fine-grained visualizations generated by GANalyze provide much clearer visual definitions.

These visualizations surface several correlates of memorability that have been overlooked by prior work, including ``object size", ``circularity", and ``colorfulness". Most past work on modeling image memorability focused on semantic attributes, such as object category (e.g., ``people" are more memorable than ``trees")~\cite{Isola2014}. By applying our approach to a class-conditional GAN, BigGAN~\cite{BigGAN}, we can restrict it to only make changes that are orthogonal to object class. This reveals more fine-grained changes that nonetheless have large effects on predicted memorability. For example, consider the cheeseburgers in Figure~\ref{fig:examples1}. Our model visualizes more memorable cheeseburgers as we move to the right. The apparent changes go well beyond semantic category -- the right-most burger is brighter, rounder, more canonical, and, we think, looks tastier.

Since our visualizations are learned based on a \emph{model} of memorability, a critical step is to verify that what we are seeing really has a causal effect on human behavior. We test this by running a behavioral experiment that measures the memorability of images generated by our GAN, and indeed we find that our manipulations have a causal effect: navigating the GAN manifold toward images that are predicted to be more memorable actually results in generating images that are measurably more memorable in the behavioral experiment.

Our contributions include the following:
\begin{itemize}
    \item Introducing GANalyze, a framework that uses GANs to provide a \emph{visual definition} of image properties, like memorability and aesthetics, that we can measure but are not easy, in words, to define. 
    \item Showing that this framework surfaces previously overlooked attributes that correlate with memorability.
    \item Demonstrating that the discovered transformations have a causal effect on memorability.
    \item Showing that GANalyze can be applied to provide visual definitions for aesthetics and emotional valence.
\end{itemize}

\subsection{Related work}
%%%%%%%%%%%%%%%%%%%%%%%%%

{\bf Generative Adversarial Networks or GANs}. 
GANs ~\cite{goodfellow} introduced a revolutionary framework to synthesize natural-looking images ~\cite{stylegan,BigGAN,sagan,proggan,BigGAN}. Among the many applications for GANs are style transfer~\cite{Zhu_2017_ICCV}, visual prediction~\cite{mathieu2015deep}, and ``sim2real" domain adaptation~\cite{graspgan}. Here, we show how they can also be applied to the problem of understanding high-level, cognitive image properties, such as memorability. 

{\bf Understanding CNN representations} 
The internal representations of a CNN can be unveiled using methods like network dissection \cite{zhou2018,bau2017,objectdetectors} including for a CNN trained on memorability \cite{Khosla_2015_ICCV}. For instance, Khosla et al. \cite{Khosla_2015_ICCV} showed that units with strong positive correlations with memorable images specialized for people, faces, body parts, etc., while those with strong negative correlations where more sensitive to large regions in landscapes scenes. Here, our framework introduces a new way of defining what memorability, and aesthetic, variability look like.

{\bf Modifying Memorability}. The memorability of an image, like faces, can be manipulated using warping techniques \cite{Khosla_2013_ICCV}. Concurrent work has also explored using a GAN for this purpose \cite{sidorov}. Another approach is a deep style transfer ~\cite{deepstyle} which taps into more artistic qualities. Now that GANs have reached a quality that is often almost indistinguishable from real images, they offer a powerful tool to synthesize images with different cognitive qualities. As shown here, our GANalyze framework successfully modified GAN-generated images across a wide range of image categories to produce a second generation of GAN realistic photos with different mnemonic qualities. 

\begin{figure*}[b]
\includegraphics[width=\textwidth]{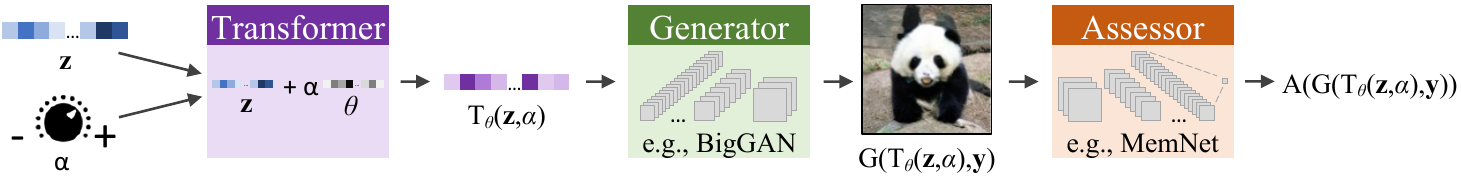}\qquad
  \caption{\textbf{Schematic of the model}. The model learns how to transform a $\mathbf{z}$ vector such that when fed to a Generator, the resulting image's property of interest changes. The transformation is achieved by the \emph{Transformer}, who moves the $\mathbf{z}$ vector along a learned direction, $\mathbf{\theta}$, in the Generator's latent space. The property of interest (e.g., memorability) is predicted by an \emph{Assessor module} (e.g., MemNet). Finally, $\alpha$ acts as knob to set the degree of change one wants to achieve in the Assessor value (e.g., MemNet score). It tells the \emph{Transformer} how far exactly to move along $\mathbf{\theta}$.}
  \label{fig:model}
\end{figure*}

%%%%%%%%%%%%%%%%%%%%%%%%%%%%%%%%%%%%%%%%%%%%%%%%%%%%%%%%%%%
\section{Model}
%%%%%%%%%%%%%%%%%%%%%%%%%%%%%%%%%%%%%%%%%%%%%%%%%%%%%%%%%%%

\subsection{Formulation}
%%%%%%%%%%%%%%%%%%%%%%%%
We start with a pretrained Generator $G$, who takes a noise vector $\mathbf{z} \in \mathbb{R}^{1\times M}$ and a one-hot class vector $\mathbf{y} \in \{0;1\}^{1\times C}$ as input and generates a photo-realistic image $G(\mathbf{z},\mathbf{y})$. Assumed is also an Assessor function $A$ that assesses an image property of interest, in this case memorability. Our goal was to learn to transform any given noise vector $\mathbf{z}$ of any class $\mathbf{y}$ such that the memorability of its resulting, generated image increases (or decreases) with a certain amount $\alpha$. The transformation is achieved by a Transformer function, who moves the input $\mathbf{z}$ along a certain direction $\mathbf{\theta} \in \mathbb{R}^{1\times M}$ in the latent space. We express the objective as:
\begin{equation}
\mathcal{L}(\theta)=\mathbb{E}_{\mathbf{z},\mathbf{y},\alpha}[(A(G(T_{\theta}(\mathbf{z},\alpha),\mathbf{y}))\\
-(A(G(\mathbf{z},\mathbf{y}))+\alpha))^2]
\end{equation}
Note that this is simply the MSE loss between the target memorability score, i.e. the seed image's score $A(G(\mathbf{z},\mathbf{y}))$ increased by $\alpha$, and the memorability score of the transformed clone image $A(G(T_{\theta}(\mathbf{z},\alpha),\mathbf{y}))$. The scalar $\alpha$ acts as a metaphorical knob with which one can use to turn up or turn down memorability. The optimizing problem is $\theta^* = \argminD_{\theta} \mathcal{L}(\theta)$.
%\begin{equation}
%\theta^* = \argminD_{\theta} \mathcal{L}(\theta)
%\end{equation}
The Transformer $T$ is defined as:
\begin{equation}
\begin{split}
T_{\theta}(\mathbf{z},\alpha) &= \mathbf{z} + \alpha \mathbf{\theta}\\
\end{split}
\end{equation}
Figure~\ref{fig:model} presents a schematic of the model. Finally, note that when $\alpha = 0$, $T$ becomes a null operation and $G(T_{\theta}(\mathbf{z},\alpha),\mathbf{y})$ then equals $G(\mathbf{z},\mathbf{y})$.

\subsection{Implementation}\label{implementation}
%%%%%%%%%%%%%%%%%%%%%%%%%%%%%%%%%%%%%%%%%%%%%%%%
For the results presented here, we used the Generator of BigGAN \cite{BigGAN}, which generates state-of-the art GAN images and is pretrained on ImageNet \cite{Imagenet}. The Assessor was implemented as MemNet \cite{Khosla_2015_ICCV}, a CNN predicting image memorability. Note, however, that training our model with different Generators or different Assessors can easily be achieved by substituting the respective modules. We discuss an Assessor for image aesthetics in Section~\ref{other}. Furthermore, we present additional results for implementations with a StyleGAN \cite{stylegan} Generator in the supplementary materials.

To train our model and find $\theta^*$, we built a training set by randomly sampling 400K $\mathbf{z}$ vectors from a standard normal distribution truncated to the range $[-2,2]$. Each $\mathbf{z}$ was accompanied by an $\alpha$ value, randomly drawn from a uniform distribution between -0.5 and 0.5, and a randomly chosen $\mathbf{y}$. We used a batch size of 4 and an Adam optimization procedure.

In view of the behavioral experiments (see Section~\ref{experiments}), we restricted the test set to 750 randomly chosen ImageNet classes and two $\mathbf{z}$ vectors per class. Each $\mathbf{z}$ vector was then paired with five different $\alpha$ values: ${[-0.2,-0.1,0,0.1,0.2]}$. Note that this includes an $\alpha$ of 0, representing the original image $G(\mathbf{z},\mathbf{y})$. Finally, the test set consisted of 1.5K sets of five images, or 7.5K test images in total.

%\section{Results}
%%%%%%%%%%%%%%%%%%%%%%%%%%%%%%%%%%%%%%%%%%%%%%%%%%%%%%%%%%%
\section{Experiments}\label{experiments}
%%%%%%%%%%%%%%%%%%%%%%%%%%%%%%%%%%%%%%%%%%%%%%%%%%%%%%%%%%%

\subsection{Model validation}
%%%%%%%%%%%%%%%%%%%%%%%%%%%%%
Did our model learn to navigate the latent space such that it can increase (or decrease) the Assessor score of the generated image with positive (or negative) $\alpha$ values?

\begin{figure}[b]
  \centering
  \includegraphics[width=0.95\textwidth]{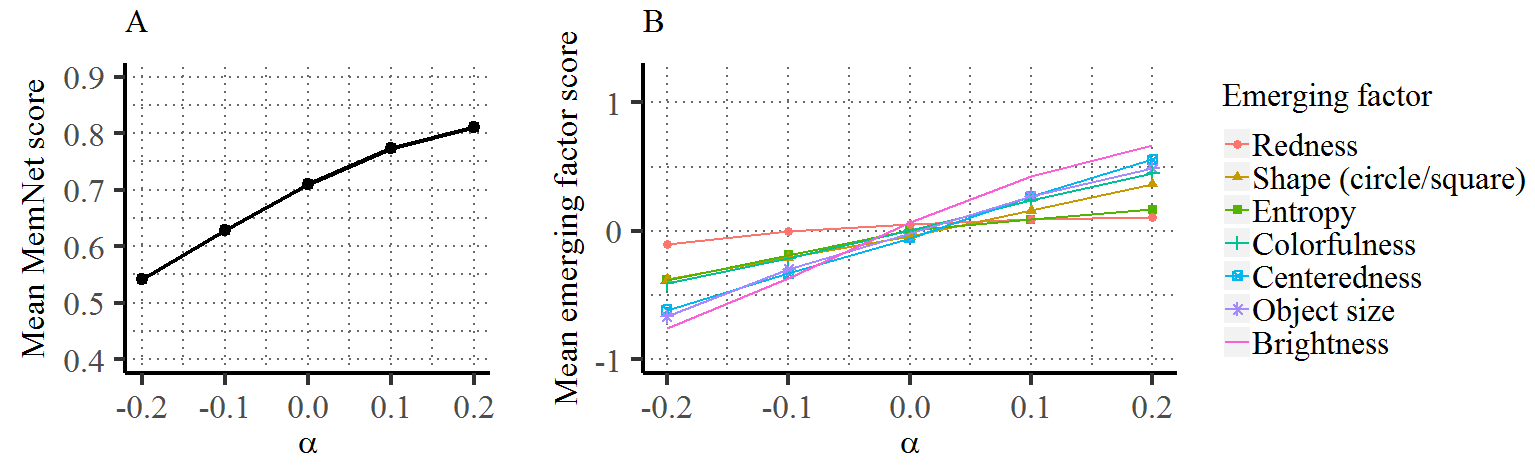}
  \caption{\textbf{Model results}. A) Graph shows the mean MemNet score across the images in every $\alpha$ condition. Our model successfully learned how to modify a GAN image to decrease (negative $\alpha$) or increase (positive $\alpha$) its MemNet score. B) List of emerging factors potentially underlying the effect observed in (A), and graph of how they change in function of $\alpha$. The factors emerged from visualizations generated by the GANalyze framework (examples shown in Figures~\ref{fig:examples1}, ~\ref{fig:sup_examples_mem_1}, and ~\ref{fig:sup_examples_mem_2}). Emerging factor scores were first normalized and then averaged per $\alpha$ condition.}
  \label{fig:memorability_graph}
\end{figure}

Figure~\ref{fig:memorability_graph}.A suggests the model learned. The mean MemNet score of test set images increases with every increment of $\alpha$. To test this formally, we fitted a linear mixed-effects regression model to the data and found a (unstandardized) slope ($\beta$) of 0.68 ($95\% CI = [0.66,0.70], p<0.001)$, confirming that the Memnet score increases significantly with $\alpha$.

\begin{figure*}[!p]
 \centering
  \includegraphics[height=0.98\dimexpr\textheight\relax]{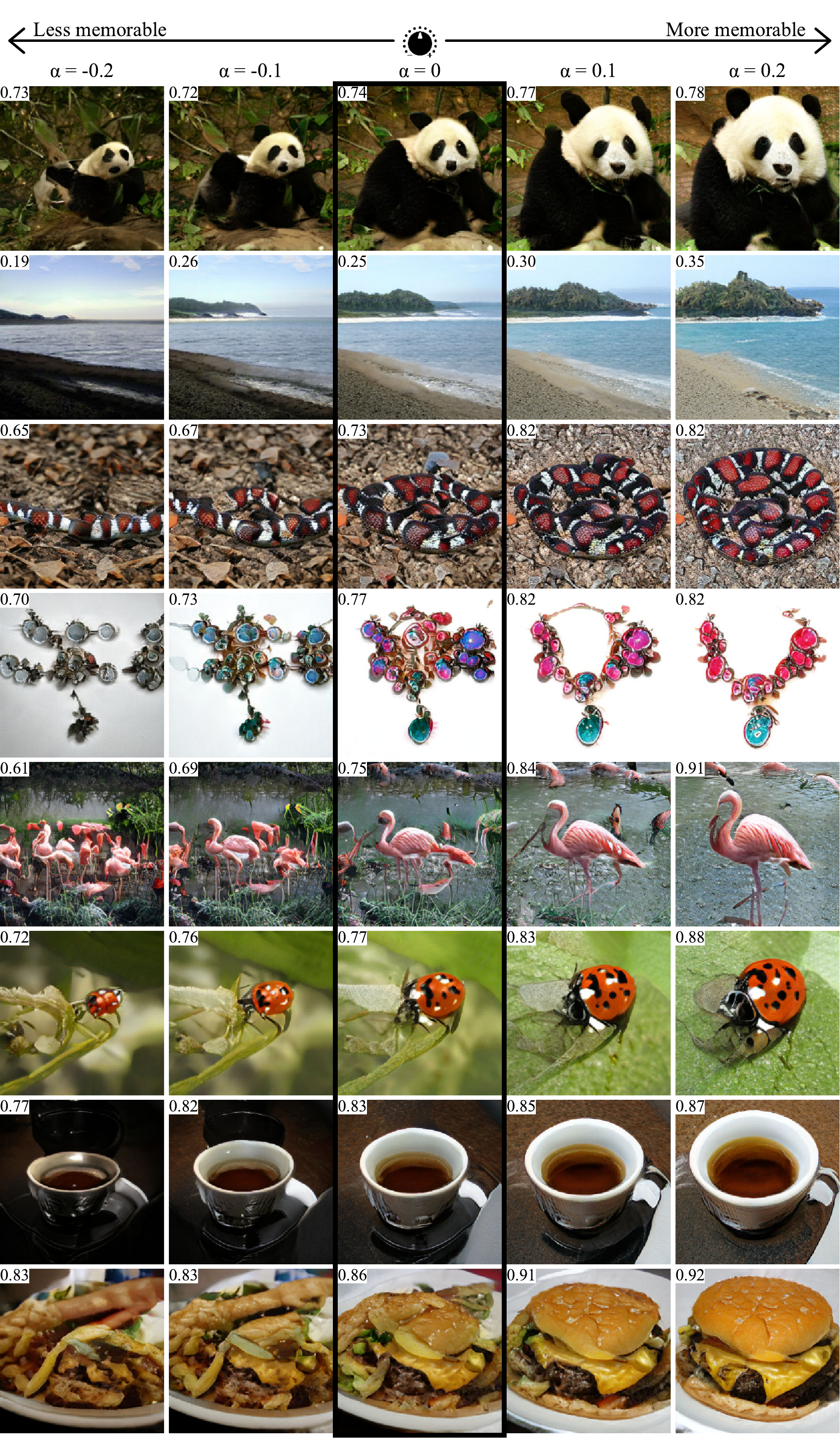}
  \caption{\textbf{Examples of generated images} along the memorability dimension. The middle column represents $G(\mathbf{z},\mathbf{y})$, the generated image serving as the original seed to create a series of clone images more or less memorable.}
  \label{fig:examples1}
\end{figure*}

\subsection{Emerging factors}\label{Emerging factors}
%%%%%%%%%%%%%%%%%%%%%%%%%%%%%
We observe that the model can successfully change the memorability of an image, given its $\mathbf{z}$ vector. Next, we ask which image factors it altered to achieve this. The answer to this question can provide further insight into what the Assessor has learned about the to-be-assessed image property, in this case what MemNet has learned about memorability. From a qualitative analysis of the test set (examples shown in Figures ~\ref{fig:examples1},  ~\ref{fig:sup_examples_mem_1}, and ~\ref{fig:sup_examples_mem_2}), a number of candidate factors stand out. 

First, MemNet assigns higher memorability scores when the \textbf{size of the object} (or animal) in the image is larger, as our model is in many cases zooming in further on the object with every increase of $\alpha$. 

Second it is \textbf{centering} the subject in the image frame.

Third, it seems to strive for \textbf{square} or \textbf{circular} shapes in classes where it is realistic to do so (e.g., snake, cheeseburger, necklace, and espresso in Figure~\ref{fig:examples1}). 

Fourth, it is often \textbf{simplifying} the image from low to high $\alpha$, by reducing the clutter and/or number of objects, such as in the cheeseburger or flamingo, or by making the background more homogeneous, as in the snake example (see Figure~\ref{fig:examples1}). 

A fifth observation is that the subject's \textbf{eyes} sometimes become more pronounced and expressive, in particular in the dog classes (see Figure~\ref{fig:teaser}). 

Sixth, one can also detect color changes between the different $\alpha$ conditions. Positive $\alpha's$ often produce \textbf{brighter} and more \textbf{colorful} images, and negative $\alpha's$ often produce darker images with dull colors. Finally, for those classes where multiple object hues can be considered realistic (e.g., the the bell pepper and the necklace in Figure~\ref{fig:teaser} and Figure~\ref{fig:examples1}), the model seems to prefer a \textbf{red} hue.

To verify our observations, we quantified the factors listed above for the images in the test set (except for "expressive eyes", which is more subjective and harder to quantify). Brightness was measured as the average pixel value after transforming the image to grayscale. For colorfulness, we used the metric proposed by \cite{colorfulness}, and for redness we computed the normalized number of red pixels. Finally, the entropy of the pixel intensity histogram was taken as proxy for simplicity. For the remaining three factors, a pretrained Mask R-CNN \cite{maskrcnn,massa2018mrcnn} was used to generate an instance-level segmentation mask of the subject. To capture object size, we calculated the difference in the mask's area (normalized number of pixels) as the step size $\alpha$ varied. To measure centeredness, we computed the deviation of the mask's centroid from the center of the frame. Finally, we calculated the length of minor and major axes of an ellipse that has the same normalized second central moments as the mask, and used their ratio as a metric of squareness. Figure~\ref{fig:memorability_graph}.B shows that the emerging factor scores  increase with $\alpha$.

\subsection{Realness}\label{Realness}
%%%%%%%%%%%%%%%%%%%%%%%%%%%%%%%%%%%%%

%\begin{figure}{r}{0.5\textwidth}
\begin{figure}[b]
%\begin{center}
\centering
\includegraphics[width=9.2cm]{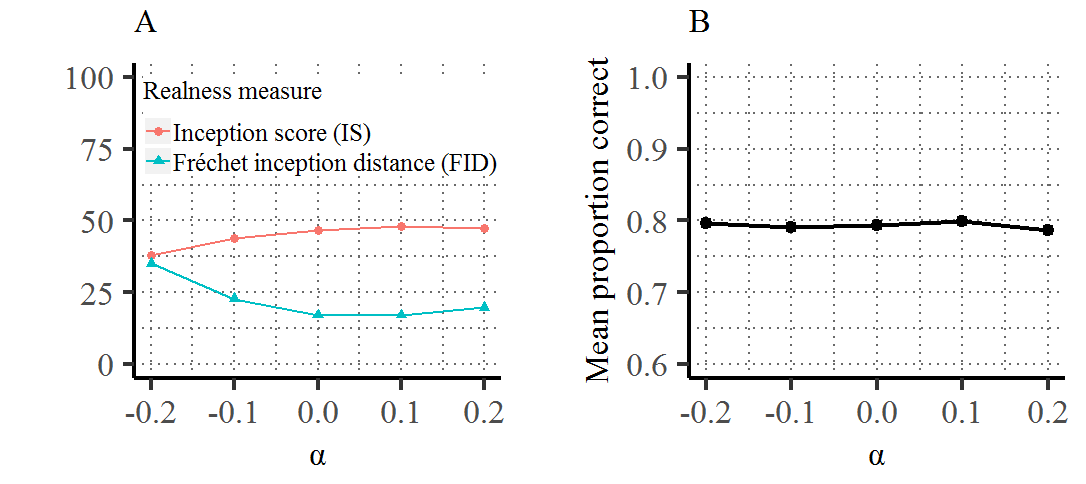}\qquad
%\end{center}
   \caption{\textbf{Realness measures as a function of $\alpha$}. A) Two popular automatic measures for evaluating the realness of a set of GAN images. Note that lower FID values indicate higher realness. B) Human fakeness discriminability, measured as the mean proportion correct in a 2AFC-task in which AMT workers had to discriminate GAN-images (fake) from real photographs.}%It was computed across all workers and all images per condition.}
  \label{fig:2afc}
\end{figure}

While BigGAN achieves state-of-the-art to generate highly realistic images, there remains a certain variability in the ``realness" of the generated images. How best to evaluate the realness of a set of GAN-images is still an open question. Below, we discuss two automatically computed realness measures and a human measure in relation to our data. We discuss an additional human measure, based on a different task, in the supplementary materials.

\subsubsection{Automatic measures}
In Figure~\ref{fig:2afc}.A, we plot two popular automatic measures in function of $\alpha$: the Frechet Inception Distance (FID) \cite{FID} and the Inception Score (IS) \cite{IS}. A first observation is that the FID is below 40 in all $\alpha$ conditions. An FID as low as 40 already corresponds to reasonably realistic images. Thus the effects of our model's modifications on memorability are not explained by making the images unrealistic. But we do observe interesting differences in FID- and IS-differences related to $\alpha$, suggesting that more memorable images have more interpretable semantics.

\subsubsection{Human measure}
In addition to the two automatic measures, we conducted an experiment to collect human realness scores. The experiment consisted of a two-alternative forced choice (2AFC) task, hosted on Amazon Mechanical Turk (AMT), in which workers had to discriminate GAN-images from real ones. Workers were shown a series of pairs, consisting of one GAN-image and one real image. They were presented side by side for a duration of 1.6 s. Once a pair had disappeared off the screen, workers pressed the j-key when they thought the GAN-image was shown on the right, or the f-key when they thought it was shown on the left. The position of the GAN-image was randomized across trials. The set of real images used in this experiment was constructed by randomly sampling 10 real ImageNet exemplars per GAN-image class. The set of GAN-images was the same as the one quantified on memorability in Section ~\ref{memory}. A GAN-image was randomly paired with one of the 10 real images belonging to the same class. Each series consisted of 100 trials, of which 20 were vigilance trials. For the vigilance trials, we generated GAN-images from $\mathbf{z}$ vectors that were sampled from the tails of a normal distribution (to make them look less real). For a worker's first series, we prepended 20 trials with feedback as practice (not included in the analyses). Workers could complete up to 17 series, but were blocked if they scored less than 65\% correct on the vigilance trials. Series that failed this criterion were also excluded from the analyses. The pay rate equaled \$0.50 per completed series. On average, each of our test images was seen by 2.76 workers, meaning 4137 data points per $\alpha$ condition.

We did not observe differences in task performance between different $\alpha$ (see Figure ~\ref{fig:2afc}.B). Indeed, a logistic mixed-effects regression fitted to the raw, binary data (correct/incorrect) did not reveal a statistically significant regression weight for $\alpha$ ($\beta=-0.08,95\%CI=[-0.33,0.18],p = 0.55$). In other words, the model's image modifications did not affect workers' ability to correctly identify the fake image, indicating that perceptually, the image clones of a seed image did not differ in realness.

%\section{Experiments}\label{experiments}
\begin{figure*}
  \includegraphics[width=480pt,height=113pt]{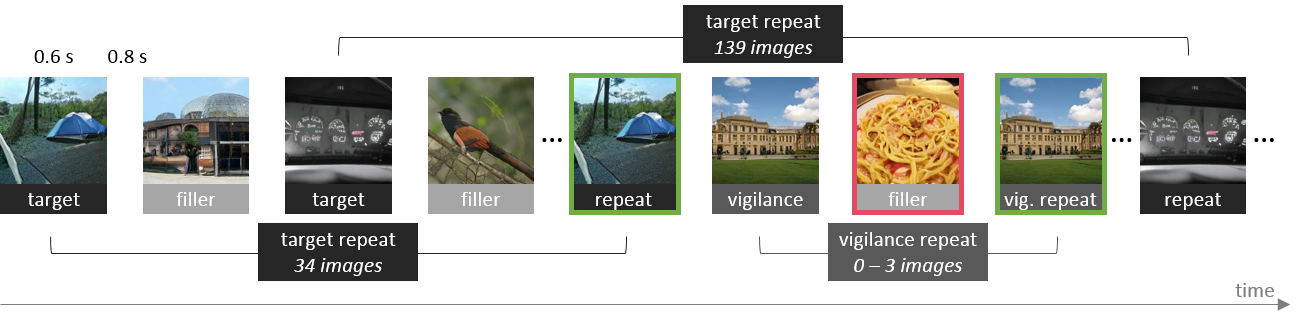}
  \caption{\textbf{Schematic of the visual memory game}. Each image is shown for 600 ms, with a blank interstimulus interval of 800 ms. Workers are asked to respond whenever they recognize a repeat of a previously shown image. For a correct response, the frame around the image briefly turns green. A red frame, on the other hand, indicates a mistake.}
  \label{fig:memory_game}
\end{figure*}
\subsection{Do our changes causally affect memory?}\label{memory}
In addition to the MemNet scores, is our model also successful at changing the probability of an image being recognized by participants in an actual memory experiment?

We tested people's memory for the images of a test set (see Section~\ref{implementation}) using a repeat-detection visual memory game, which was hosted on AMT (see Figure~\ref{fig:memory_game}). \cite{Isola2014,Khosla_2015_ICCV}. AMT workers watched a series of one image at the time and had to press a key whenever they saw a repeat of a previously shown image. Each series consisted of 215 images, shown each for 600 ms with a blank interstimulus interval of 800 ms. Sixty images were targets, sampled from our test set, and repeated after 34 to 139 intervening images. The remaining images were either filler or vigilance images and were sampled from a separate set. This set was created with 10 $\mathbf{z}$ vectors per class and the same five $\alpha$ values as the test set: $[-0.2,-0.1,0,0.1,0.2]$, making a total of 37.5K images. Filler images were only presented once and ensured spacing between a target and its repeat. Vigilance images were presented twice, with 0 to 3 intervening images in-between the two presentations. The vigilance repeats constituted easy trials to keep workers attentive. Care was taken to ensure that a worker never saw more than one $G(T_{\theta}(\mathbf{z},\alpha),\mathbf{y})$ for a given $\mathbf{z}$. Workers could complete up to 25 series, but were blocked if they missed more than 55\% of the vigilance repeats in a series or made more than 30\% false positives. Series that failed this were excluded from the analyses. The pay rate was \$0.50 per completed series. On average, a test image was seen by 3.16 workers, with 4740 data points per $\alpha$ condition.

\begin{figure}[!h]
  \centering
  \includegraphics[width=12.45cm]{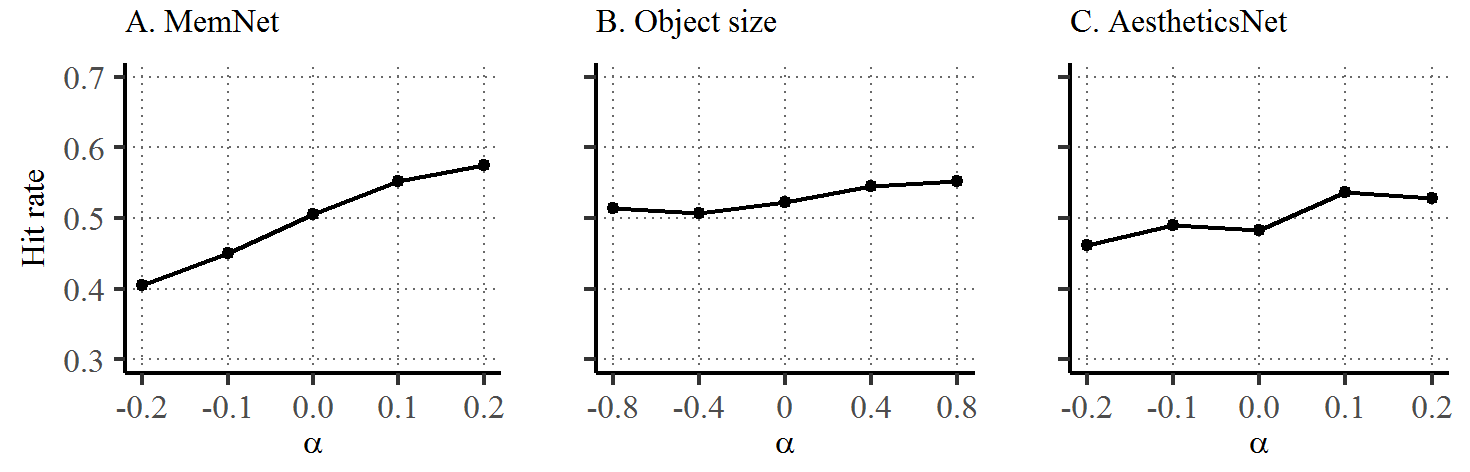}
  \caption{\textbf{Human memory performance} for images modified according to different Assessors: A) MemNet, B) Object size and C) AestheticsNet. Performance is measured as the hit rate across all images and workers in the memory game for each property.} 

  \label{fig:experiments_plot}
\end{figure}

Workers could either recognize a repeated test image (hit, 1), or miss it (miss, 0). % meaning they forgot that they had seen the image before. 
Figure~\ref{fig:experiments_plot}.A shows the hit rate across all images and workers. The hit rate increases with every step of $\alpha$. Fitting a logistic mixed-effects regression model to the raw, binary data (hit/miss), we found that the predicted log odds of image being recognized increase with 0.19 for an increase in $\alpha$ of 0.01 ($\beta = 1.92, 95\%CI=[1.71-2.12],p<0.001$). This shows that our model can successfully navigate the BigGAN latent space in order to make an image more (or less) memorable to humans. 

\subsubsection{Emerging factors}
Given human memory data for images modified for memorability, we evaluate how the images' emerging factor scores relate to their likelihood of being recognized. We fitted mixed-effects logistic regression models, each with a different emerging factor as the predictor, see Table ~\ref{table:emerging_factors}. Except for entropy, all the emerging factors show a significant, positive relation to the likelihood of a hit in the memory game, but none fit the data as well as the model's $\alpha$.  This indicates that a single emerging factor is not enough to fully explain the effect observed in Figure ~\ref{fig:experiments_plot}.A. Note that the emerging factor results are correlational and the factors are intercorrelated. This makes it hard to draw conclusions about which individual factors truly causally affect human memory performance. As an example of how this can be addressed within the GANalyze framework, we conducted an experiment focusing on the effect of one salient emerging factor: \textbf{object size}. As seen in Figure~\ref{fig:examples1}, more memorable images tend to center and enlarge the object class.  

\begin {table}
\begin{center}
\begin{tabular}{
  l|
  S[table-format = -1.2, table-space-text-post = $^{***}$]|
  >{{[}} % Add square bracket before column
    S[table-format = -1.2,table-space-text-pre={[}]
    @{,\,} % Add comma and thin-space between the columns
    S[table-format = -1.2,table-space-text-post={]}]
    <{{]}}| % Add square bracket after column
  S[table-format = <2.2]|
  S[table-format = 1.3]}
\toprule
{Factor} & {Log Odds} & \multicolumn{2}{c|}{CI} & {$p$} & {Tjur's D}\\
\midrule
Brightness & 0.28 & 0.24 & 0.32 & <0.001 &0.066\\ 
Centeredness & 0.24 & 0.19 & 0.29 & <0.001 &0.059\\ 
Colorfulness & 0.17 & 0.14 & 0.21 & <0.001 &0.054\\
Entropy& 0.03 & -0.04 & 0.10 & 0.441 & 0.062\\
Redness& 0.06 & 0.00 & 0.12 & 0.042 & 0.055\\
Shape & 0.19 & 0.14 & 0.24 & <0.001&0.060\\ 
Object size & 0.32 & 0.27 & 0.37 & < 0.001 &0.050 \\
\midrule
$\alpha$ & 1.92 & 1.71 & 2.12 & <0.001 & 0.074\\
\bottomrule
\end{tabular}
\vspace{5pt}
\caption{
\small{\textbf{Relation between emerging factors and human memory performance.} We show the output of logistic mixed-effects regressions. From left to right: the regression weight, the confidence interval (CI) for that weight, the $p$-value for statistical significance, and Tjur's coefficient of discrimination (D), being the regression model's goodness of fit \cite{tjur}. The emerging factor values were normalized before running the regression models.}
}
\end{center}
\label{table:emerging_factors}
\end{table}

\begin{figure}
 \centering
  \includegraphics[width=0.63\textwidth]{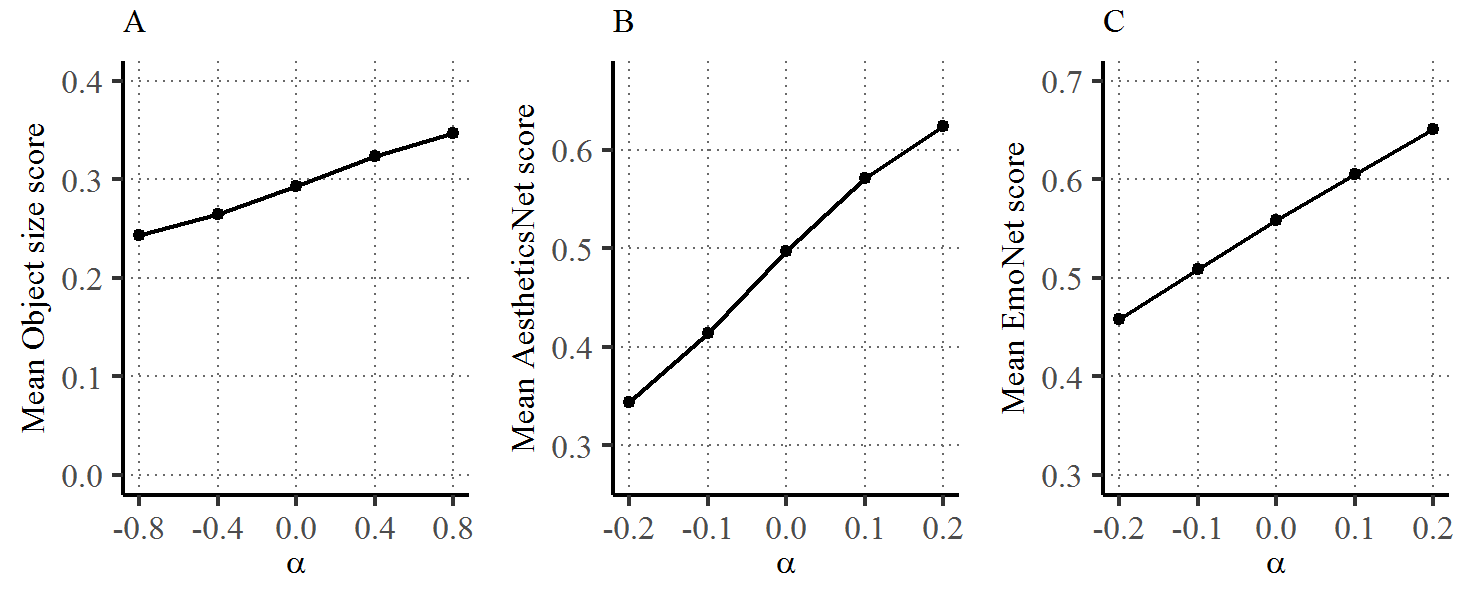}
  \caption{\textbf{Model results for additional Assessors}. A) Graph shows the mean Object size Assessor score across images in every $\alpha$ condition. A different set of $\alpha$ values was chosen to qualitatively match the degree object size changes achieved by the model trained with the MemNet Assessor. B) Graph shows the mean AestheticsNet score across the images in every $\alpha$ condition. C) Graph shows the mean EmoNet score across the images in every $\alpha$ condition}
  \label{fig:aesthetic_validation}
\end{figure}

We trained a version of our model with an Object size Assessor, instead of the MemNet Assessor. This is the same Object size Assessor used to quantify the object size in the images modified according to MemNet (e.g., for the results in Figure ~\ref{fig:memorability_graph}.B), now teaching the Transformer to perform ``enlarging" modifications. After training with 161750 $\mathbf{z}$ vectors, we generated a test set as described in Section ~\ref{implementation}, except that we used a different set of $\alpha$'s: $[-0.8,-0.4,0,0.4,0.8]$. We chose these values to qualitatively match the degree of object size changes achieved by the MemNet version of the model. Figure ~\ref{fig:aesthetic_validation}.A visualizes the results achieved on the test set. The model successfully enlarges the object with increasing alpha's, as confirmed by a linear mixed-effects regression analysis ($\beta=0.07,95\%CI=[0.06,0.07],p <0.001$). Figure~\ref{fig:comparison} shows example images generated by that model, after having been trained with 161750 $\mathbf{z}$ vectors. A comparison with images modified according to MemNet suggests that the latter model was doing more than just enlarging the object.

To study how the new size modifications affect memorability, we generated a new set of images (7.5K targets, 37.5K fillers) with $\alpha$'s $[-0.8,-0.4,0,0.4,0.8]$. The new images were then quantified using the visual memory game (on average 2.36 data points per image and 3540 per $\alpha$ condition). Figure ~\ref{fig:experiments_plot}.B shows the results. Memory performance increases with $\alpha$, as confirmed by a logistic mixed-effects analysis ($\beta = 0.11, 95\%CI=[0.06,0.18],p<0.001$, although  mostly for positive $\alpha$ values.

%%%%%%%%%%%%%%%%%%%%%%%%%%%%%%%%%%%%%%%%%%%%%%%%%%%%%%%%%%%
\section{Other properties}\label{other}
%%%%%%%%%%%%%%%%%%%%%%%%%%%%%%%%%%%%%%%%%%%%%%%%%%%%%%%%%%%
As mentioned in Section~\ref{implementation}, the proposed method can be applied to other image properties, simply by substituting the Assessor module. To show our framework can generalize, we trained a model for aesthetics, using Kong et al's ~\cite{Kong_et_al_2016} CNN (hereinafter referred to as AestheticsNet) as the Assessor. In addition, we also trained a model for emotional valence. Emotional valence refers to the extent to which the emotions evoked by an image are experienced as positive (or negative). For this property, we trained our own Assessor by fine-tuning a ResNet50 model ~\cite{resnet}, pretrained on the Moments database ~\cite{monfort2019moments}, to the Cornell Emotion6 Image Database ~\cite{emotion6}. We refer to this Assessor as EmoNet. Finally, we generated a test set for each of the two new models, like we did for the memorability model.

\begin{figure}[tbp]
 \centering
  \includegraphics[width=0.37\textwidth]{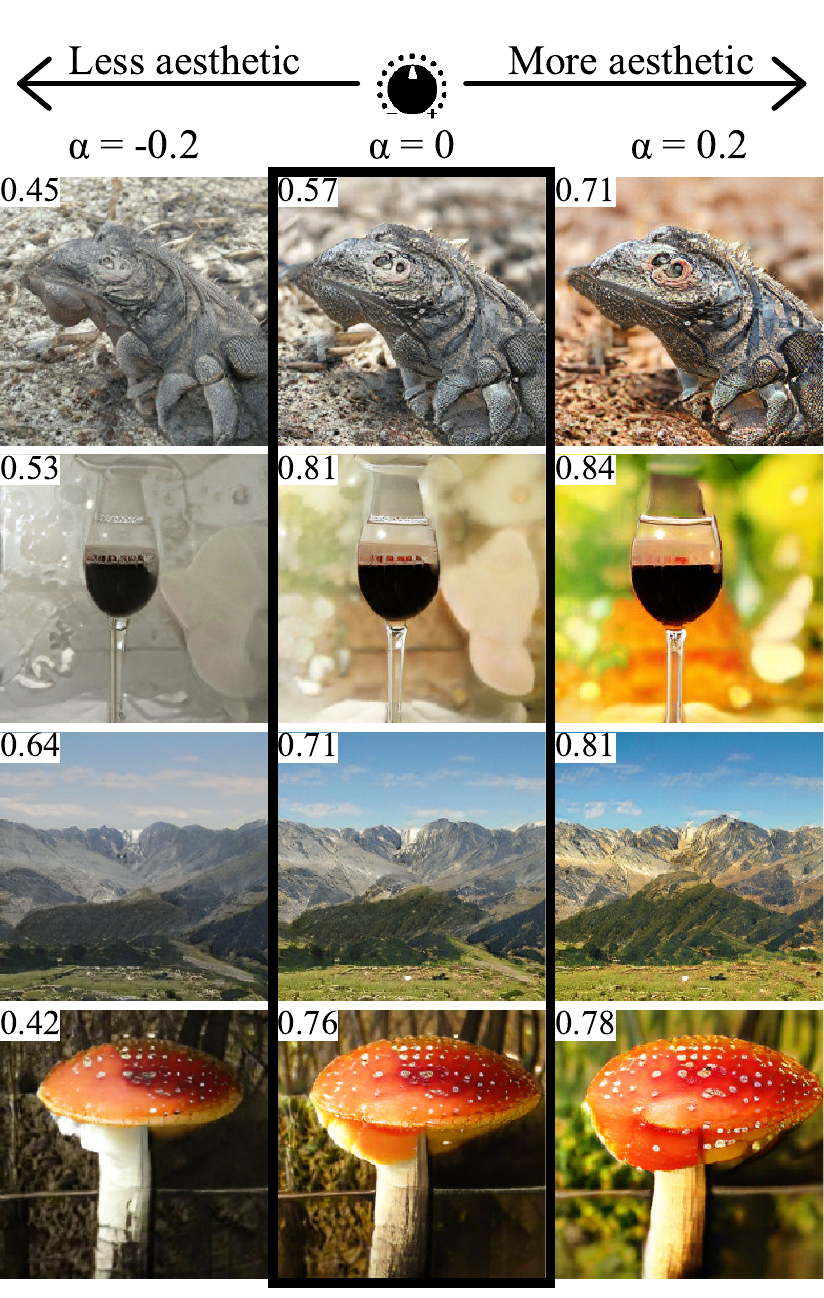}
  \qquad
  \qquad
  \includegraphics[width=0.37\textwidth]{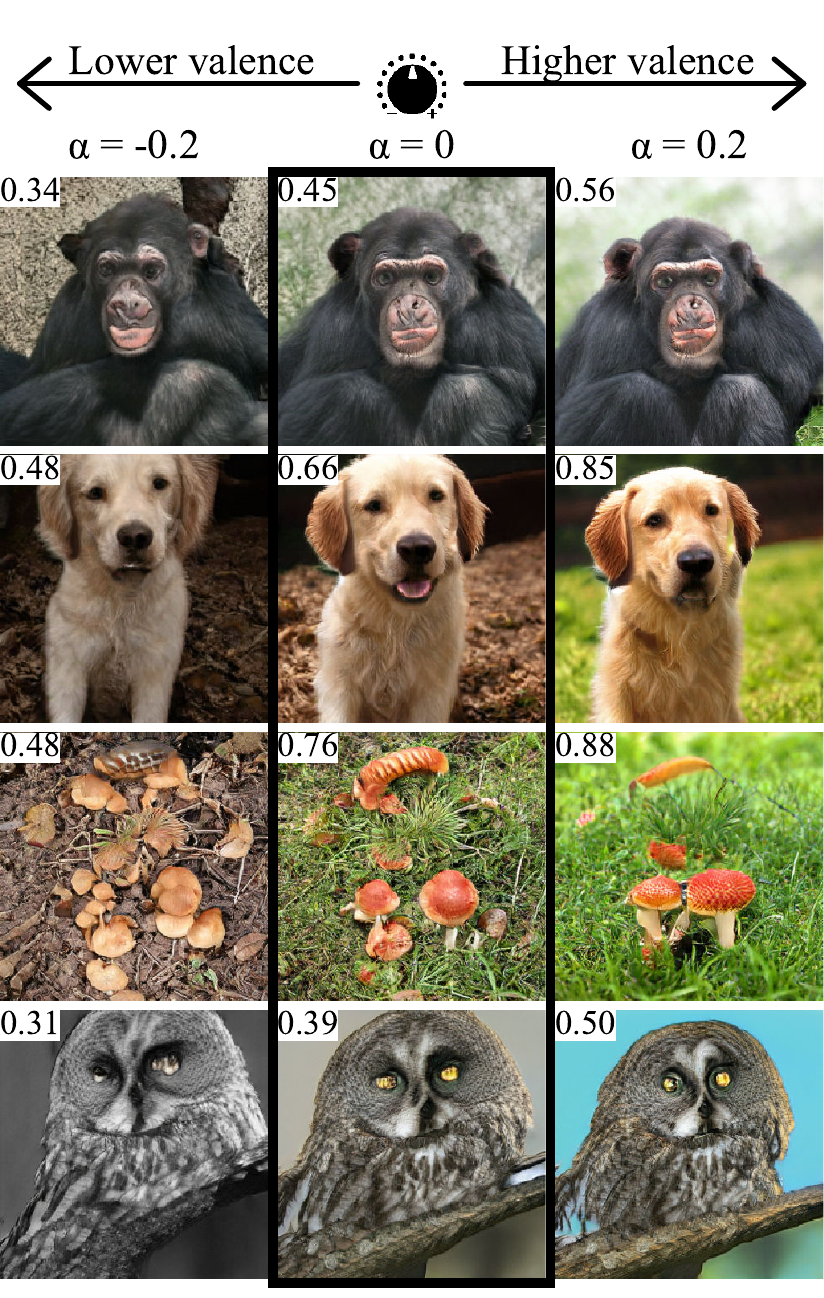}
  \caption{\textbf{Examples of generated images} along the aesthetics dimension (left) and the emotional valence dimension (right). Each middle column represents $G(\mathbf{z},\mathbf{y})$, the generated image serving as the original seed to create a series of clone images scoring higher or lower on the respective dimension according to the Assessor. The images' Assessor scores are presented in their top left corner.}
  \label{fig:examples_aes1}
\end{figure}

Figure~\ref{fig:aesthetic_validation}.B shows the average AestheticsNet scores per $\alpha$ condition. The scores significantly increase with $\alpha$, as evidenced by the results of a linear mixed-effects regression ($\beta = 0.72, 95\% CI = [0.70,0.74],p<0.001$). We can successfully train the model to increase (or decrease) an image's aesthetic score as shown in Figure~\ref{fig:examples_aes1} (left) and Figure~\ref{fig:sup_examples_aes_1}. Similarly, Figure~\ref{fig:aesthetic_validation}.C shows the average EmoNet scores per $\alpha$ condition. Here too, the scores significantly increase with $\alpha$ ($\beta = 0.44, 95\% CI = [0.43,0.45],p<0.001$). Example visualizations generated by this model are presented in Figure~\ref{fig:examples_aes1} (right) and Figure~\ref{fig:sup_examples_emo_1}.

\begin{figure}[tbp]
\centering
  \includegraphics[width=0.9\textwidth]{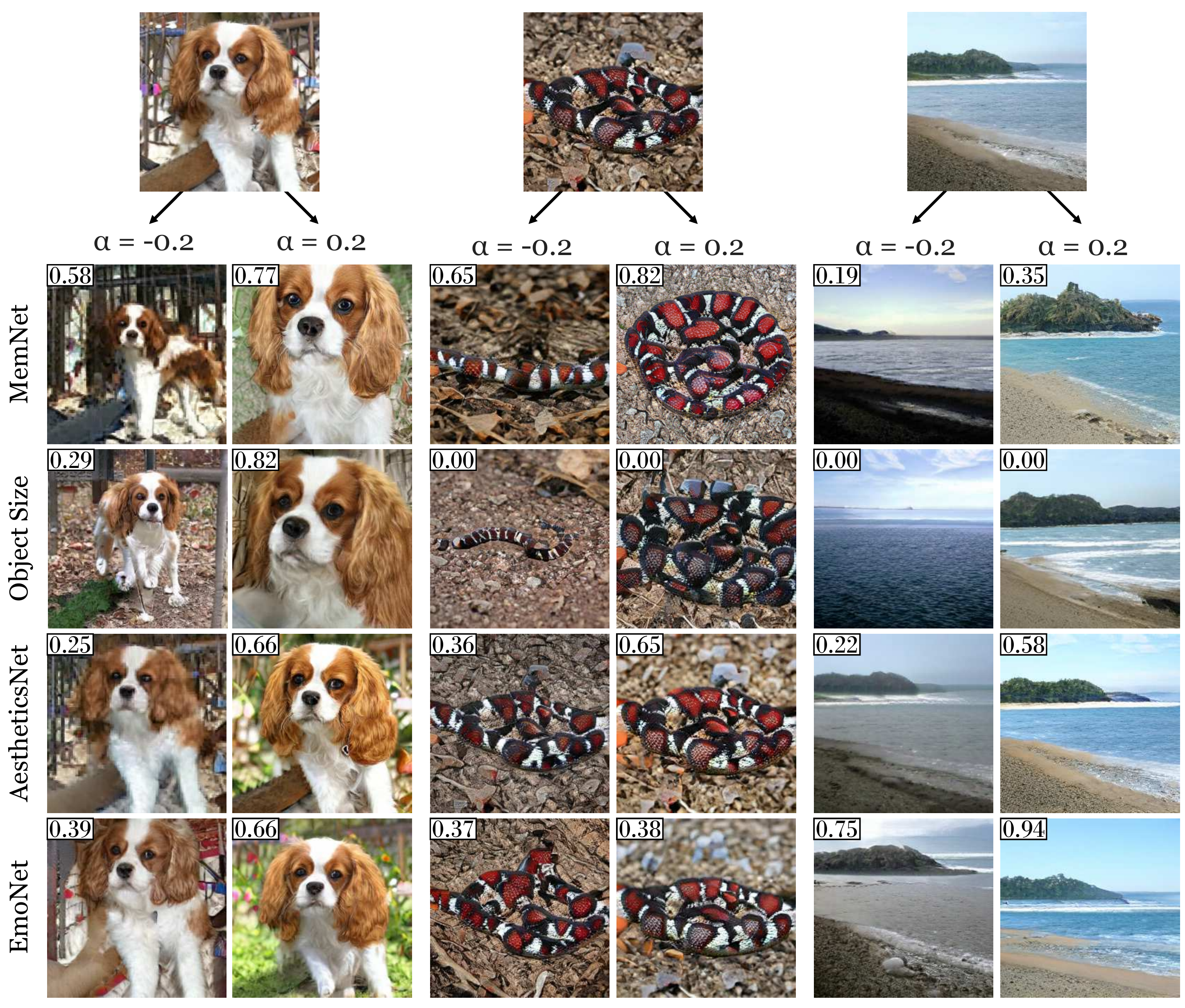}
  \caption{\textbf{Comparison of examples generated according to different Assessors.} The top row represents $G(\mathbf{z},\mathbf{y})$, the generated image serving as the original seed to create a series of images with a higher or lower Assessor value. The respective Assessor values are indicated in the top left corner. Note that for object size, we used a different $\alpha$ range: \{-0.8,0.8\}.}
  \label{fig:comparison}
\end{figure}

Based on a qualitative inspection of such visualizations, we observed that the aesthetics model is modifying factors like depth of field, color palette, and lighting, suggesting that the AestheticsNet is sensitive to those factors. Indeed, the architecture of the AestheticsNet includes attribute-adaptive layers to predict these factors, now highlighted by our visualizations. The emotional valence model often averts the subject's gaze away from the "camera" when decreasing valence. To increase valence, it often makes images more colorful, introduces bokeh, and makes the skies more blue in landscape images. Finally, the teddy bear in Figure~\ref{fig:teaser} (right) seems to smile more. Interestingly, the model makes different modifications for every property (see Figure~\ref{fig:comparison}), suggesting that what makes an image memorable is different from what makes it aesthetically pleasing or more positive in its emotional valence.

A final question we asked is whether an image modified to become more (less) aesthetic also becomes more (less) memorable? To test this, we quantified the images of the aesthetic test set on memorability by presenting them to workers in the visual memory game (we collected 1.54 data points per image and 2306 data points per $\alpha$ condition). Figure~\ref{fig:experiments_plot}.C shows the human memory performance in function of an $\alpha$ that is tuning aesthetics. A logistic mixed-effects regression revealed that with an 0.1 increase in the aesthetics $\alpha$, the predicted log odds of an image being recognized increase with 0.07 ($\beta = 0.72, 95\% CI = [0.44,1.00],p<0.001$). While modifying an image to make it more aesthetic does increase its memorability, the effect is rather small, suggesting that memorability is more than only aesthetics and that our model was right to modify memorability and aesthetics in different ways.

%%%%%%%%%%%%%%%%%%%%%%%%%%%%%%%%%%%%%%%%%%%%%%%%%%%%%%%%%%%
\section{Conclusion}
%%%%%%%%%%%%%%%%%%%%%%%%%%%%%%%%%%%%%%%%%%%%%%%%%%%%%%%%%%%
We introduce GANalyze, a framework that shows how a GAN-based model can be used to visualize what another model (i.e. CNN as an Assessor) has learned about its target image property. Here we applied it to memorability, yielding a kind of ``visual definition" of this high-level cognitive property, where we visualize what it looks like for an image to become more or less memorable. These visualizations surface multiple candidate features that may help explain why we remember what we do. Importantly, our framework can also be generalized to other image properties, such as aesthetics or emotional valence: by replacing the Assessor module, the framework allows us to explore the visual definition for any property we can model as a differentiable function of the image. We validated that our model successfully modified GAN images to become more (or less) memorable via a behavioral human memory experiment on manipulated images.

GANalyze's intended use is to contribute to the scientific understanding of otherwise hard to define cognitive properties. Note that this was achieved by modifying images for which the encoding into the latent space of the GAN was given. In other words, it is currently only possible to modify seed images that are GAN-images themselves, not user-supplied, real images. However, should advances in the field lead to an encoder network, this would become possible and it would open applications in graphics and education, for example, where selected images can be made more memorable. One should also be wary, though, of potential misuse, especially when applied to images of people or faces. Note that the BigGAN~\cite{BigGAN} generator used here was trained on ImageNet categories~\cite{Imagenet} which only occasionally include people, and that it does not allow to render realistically looking people. Nevertheless, with generative models yielding ever more realistic output, an increasingly important challenge in the field is to develop powerful detection methods to allow us to reliably distinguish generated, fake images from real ones~\cite{washington_post}\cite{technology_review}\cite{hill}.

%%%%%%%%%%%%%%%%%%%%%%%%%%%%%%%%%%%%%%%%%%%%%%%%%%%%%%%%%%%
\section{Acknowledgments}
%%%%%%%%%%%%%%%%%%%%%%%%%%%%%%%%%%%%%%%%%%%%%%%%%%%%%%%%%%%
This work was partly funded by NSF award 1532591 in Neural and Cognitive Systems (to A.O), by a fellowship (Grant 1108116N) and a travel grant (Grant V4.085.18N) awarded to Lore Goetschalckx by the Research Foundation - Flanders (FWO).

\bibliographystyle{unsrt}  
\bibliography{references}  %%% Remove comment to use the external .bib file (using bibtex).

\beginsupplement
\begin{figure*}[!p]
 \centering
  \includegraphics[height=0.98\dimexpr\textheight\relax]{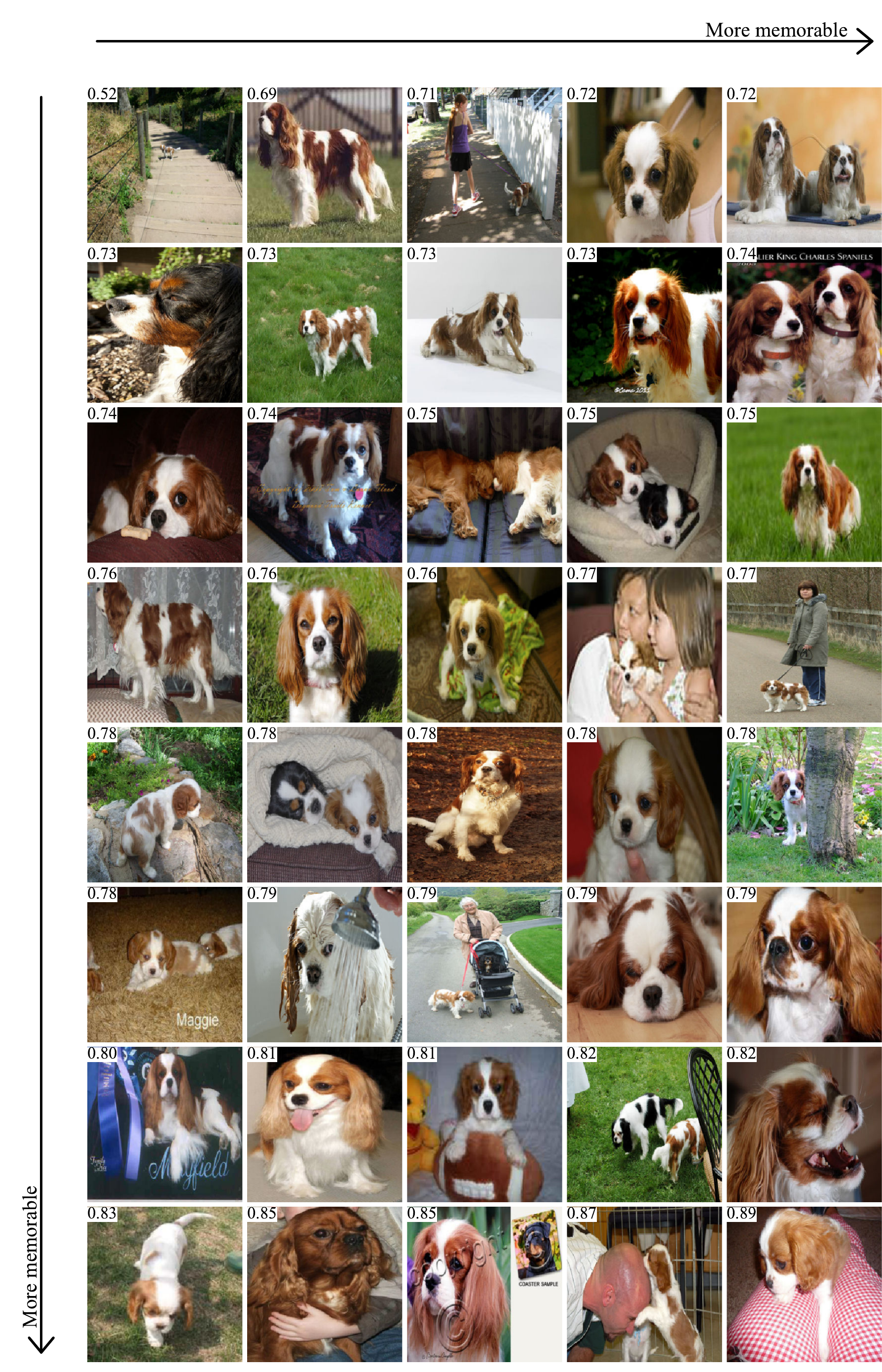}
  \caption{\textbf{Real images sorted on their MemNet score.} As opposed to GANalyze, this is a non-parametric way of visualizing what it looks like for images to become more memorable.}
  \label{fig:nonparametric}
\end{figure*}

\begin{figure*}[!p]
 \centering
  \includegraphics[height=0.98\dimexpr\textheight\relax]{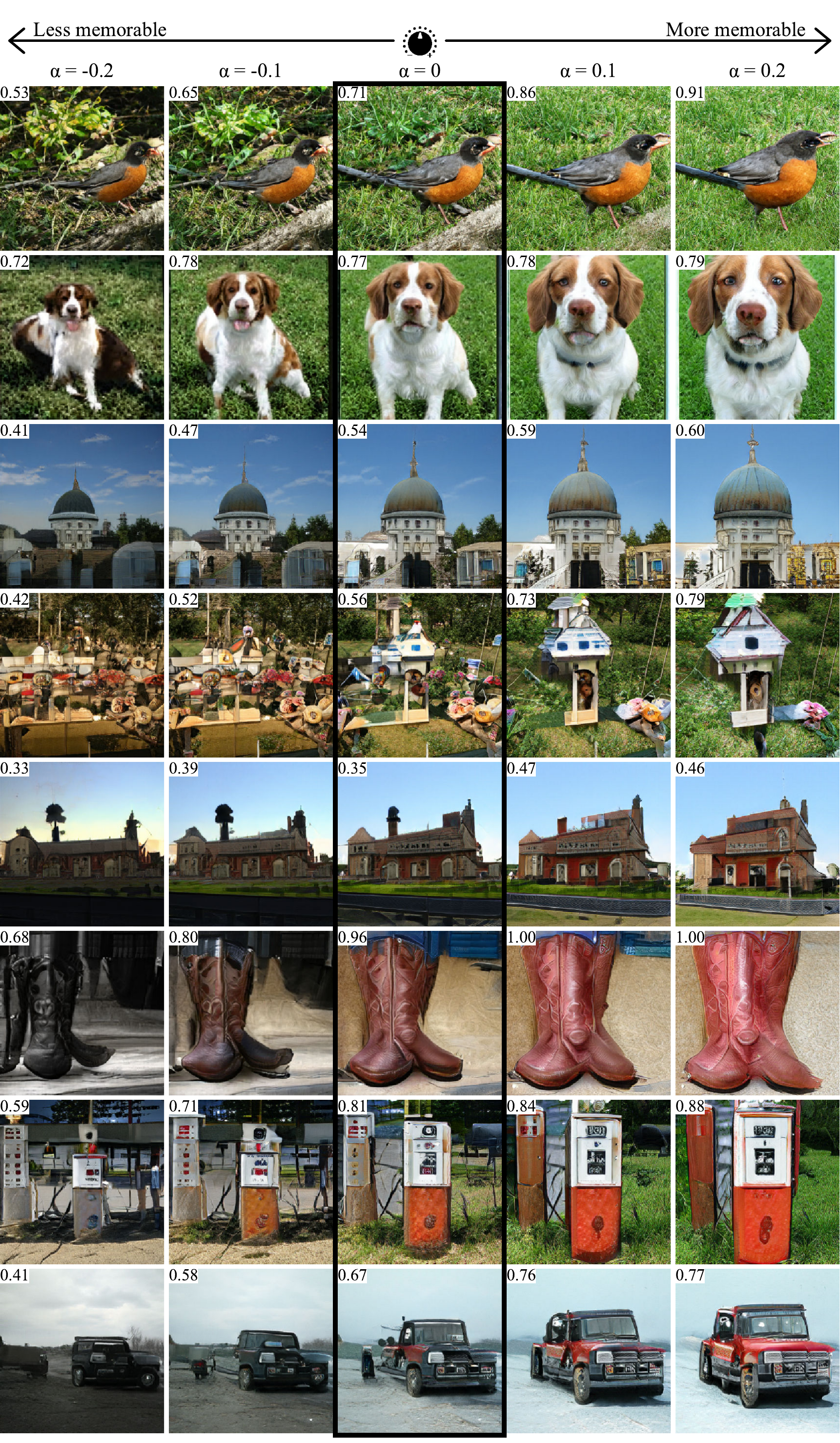}
  \caption{\textbf{More examples of generated images} along the memorability dimension. The middle column represents $G(\mathbf{z},\mathbf{y})$, the generated image serving as the original seed to create a series of clone images more or less memorable.}
  \label{fig:sup_examples_mem_1}
\end{figure*}

\begin{figure*}[!p]
 \centering
  \includegraphics[height=0.98\dimexpr\textheight\relax]{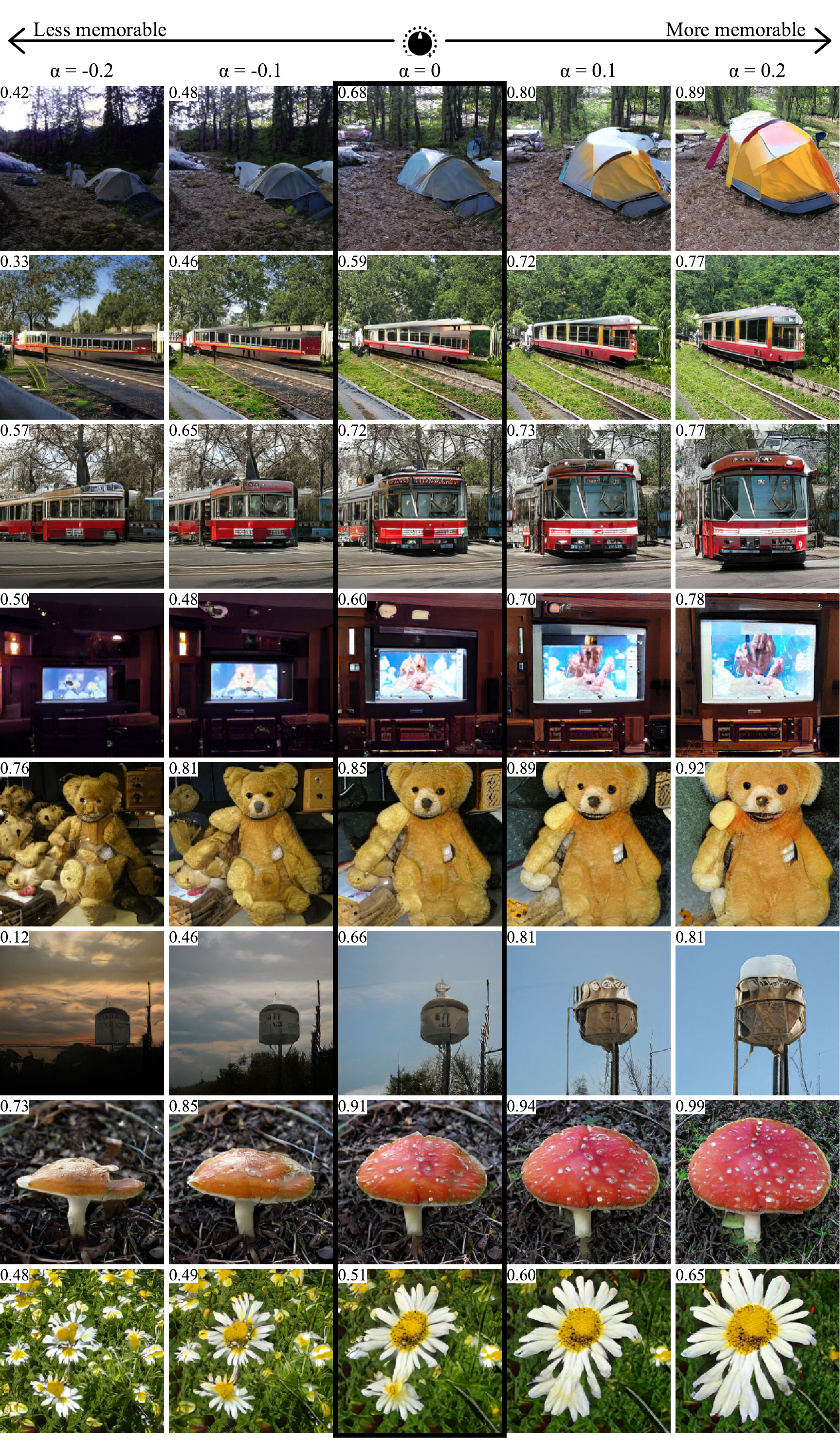}
  \caption{\textbf{More examples of generated images} along the memorability dimension. The middle column represents $G(\mathbf{z},\mathbf{y})$, the generated image serving as the original seed to create a series of clone images more or less memorable.}
  \label{fig:sup_examples_mem_2}
\end{figure*}

\begin{figure*}[!p]
 \centering
  \includegraphics[height=0.98\dimexpr\textheight\relax]{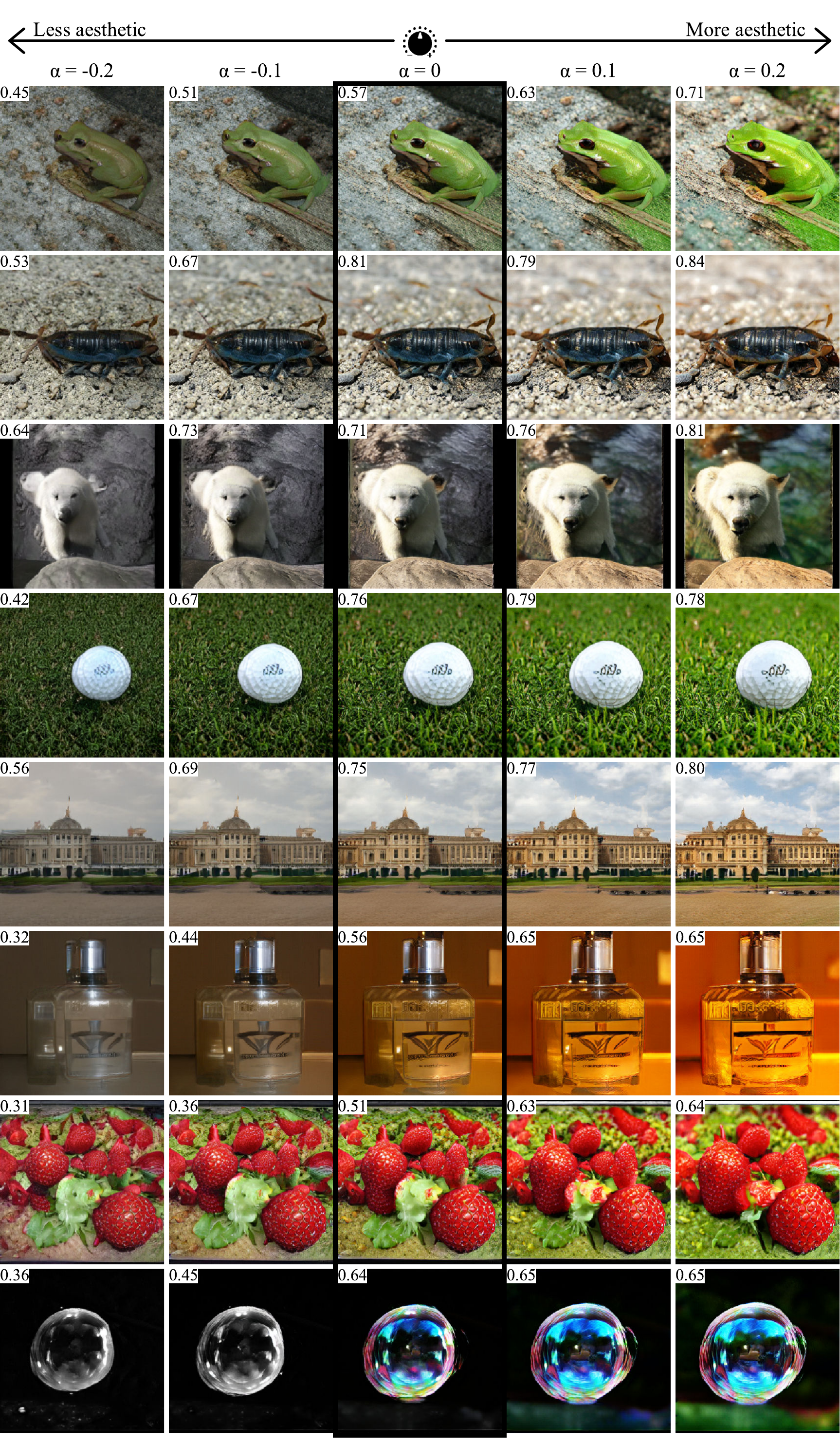}
  \caption{\textbf{More examples of generated images} along the aesthetics dimension. The middle column represents $G(\mathbf{z},\mathbf{y})$, the generated image serving as the original seed to create a series of clone images more or less aesthetic.}
  \label{fig:sup_examples_aes_1}
\end{figure*}

\begin{figure*}[!p]
 \centering
  \includegraphics[height=0.98\dimexpr\textheight\relax]{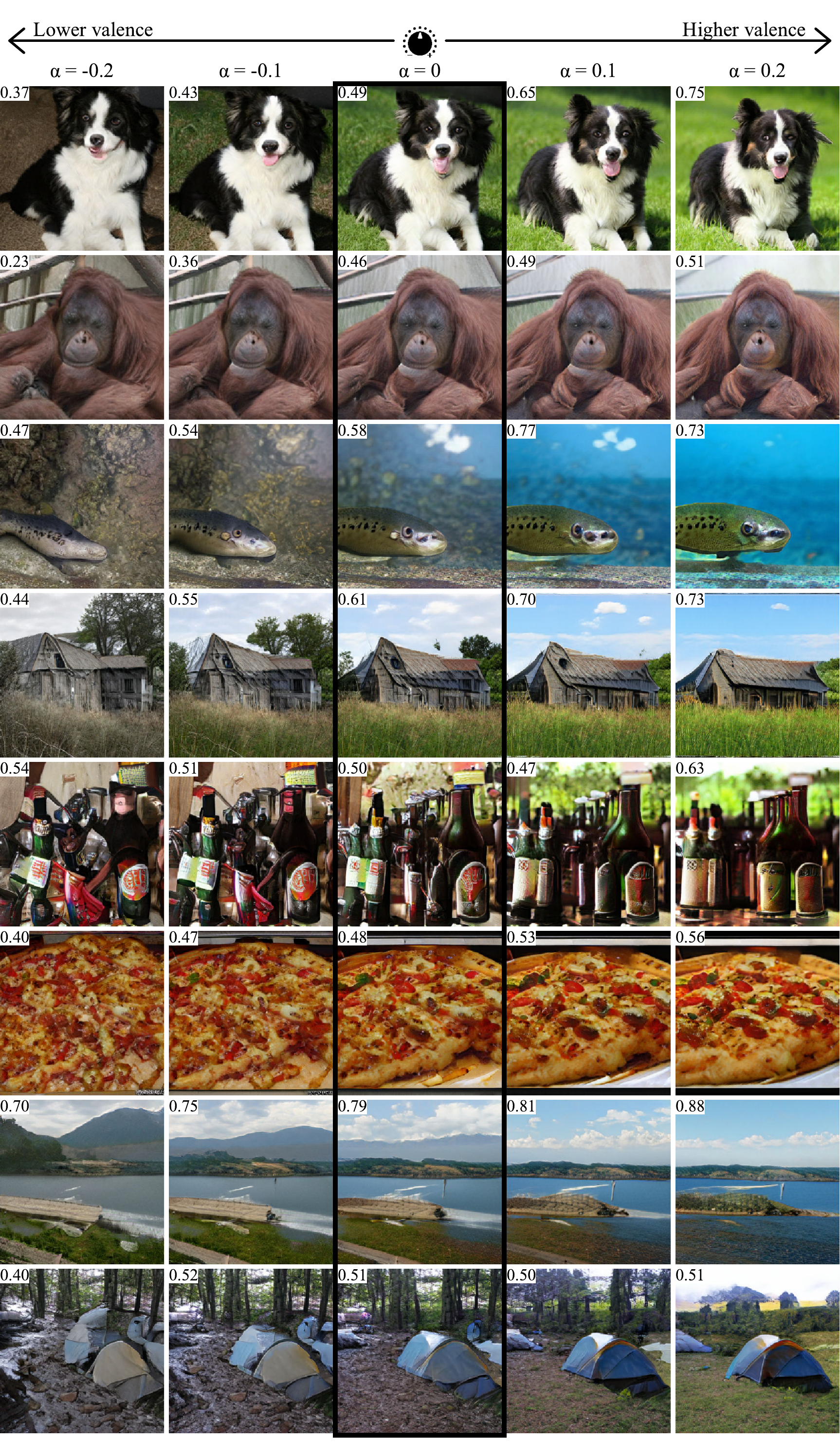}
  \caption{\textbf{More examples of generated images} along the emotional valence dimension. The middle column represents $G(\mathbf{z},\mathbf{y})$, the generated image serving as the original seed to create a series of clone images higher or lower in emotional valence.}
  \label{fig:sup_examples_emo_1}
\end{figure*}

\end{document}